\documentclass{article}

\usepackage[final]{nips_2016}

\usepackage[utf8]{inputenc} 
\usepackage[T1]{fontenc}    
\usepackage{hyperref}       
\usepackage{url}            
\usepackage{booktabs}       
\usepackage{amsfonts}       
\usepackage{microtype}      

\usepackage{amssymb, amsmath}
\usepackage{epsfig}
\usepackage{array}
\usepackage{ifthen}
\usepackage{color}
\usepackage{fancyhdr}
\usepackage{graphicx}
\usepackage{mathtools}
\usepackage{csquotes}
\usepackage{xcolor}
\usepackage{multirow}
\newcommand\crule[3][black]{\textcolor{#1}{\rule{#2}{#3}}}

\newcommand{\E}{\textbf{E}}

\newcommand{\argmax}{\text{argmax}}

\newcommand{\comm}[1]{}

\definecolor{color1}{RGB}{128,13,13}
\definecolor{color2}{RGB}{70,128,13}
\definecolor{color3}{RGB}{13,128,128}
\definecolor{color4}{RGB}{70,13,128}

\title{How many faces can be recognized? Performance extrapolation for
  multi-class classification}

\author{
  Charles Y.~Zheng \\
  Department of Statistics\\
  Stanford University\\
  Stanford, CA 94305 \\
  \texttt{snarles@stanford.edu} \\
  \And
  Rakesh ~Achanta \\
  Department of Statistics\\
  Stanford University\\
  Stanford, CA 94305 \\
  \texttt{rakesha@stanford.edu} \\
  \And
  Yuval ~Benjamini \\
  Department of Statistics \\
  Hebrew University\\
  Jerusalem, Israel\\
  \texttt{yuval.benjamini@mail.huji.ac.il}
}

\begin{document}

\maketitle

\begin{abstract}
The difficulty of multi-class classification generally increases with
the number of classes.  Using data from a subset of the classes, 
can we predict how well a classifier will scale with an
increased number of classes?  Under the assumption that the classes
are sampled exchangeably, and under the assumption that
the classifier is generative (e.g. QDA or Naive Bayes), we show that the expected accuracy
when the classifier is trained on $k$ classes is the $k-1$st moment
of a \emph{conditional accuracy distribution}, which can be estimated from data.
This provides the theoretical foundation for performance extrapolation based on pseudolikelihood, 
unbiased estimation, and high-dimensional asymptotics.
We investigate the robustness of our methods to non-generative classifiers
in simulations and one optical character recognition example.
\end{abstract}

\section{Introduction}

In multi-class classification, one observes pairs $(z, y)$ where $y \in \mathcal{Y} \subset \mathbb{R}^p$ are feature vectors,
and $z$ are unknown labels, which lie in a countable label set $\mathcal{Z}$.  The goal is to construct a classification rule for
predicting the label of a new data point; generally, the classification rule $h: \mathcal{Y} \to \mathcal{Z}$
is learned from previously observed data points.  In many applications of multi-class classification,
such as face recognition or image recognition, the space of potential labels is practically infinite.
In such a setting, one might consider a sequence of classification problems on finite label subsets $\mathcal{Z}_1 \subset \cdots \subset \mathcal{Z}_K$, where in the $i$-th problem, one constructs the classification rule $h^{(i)}:\mathcal{Y} \to \mathcal{Z}_i$.
Supposing that $(Z, Y)$ have a joint distribution, define the accuracy for the $i$-th problem as
\[
\text{acc}^{(i)} = \Pr[h^{(i)}(Y) = Z|Z \in \mathcal{Z}_i].
\]
Using data from only $\mathcal{Z}_k$, can one predict the accuracy achieved on the larger label set $\mathcal{Z}_K$, with $K> k$?  This is the problem of \emph{performance extrapolation}.

A practical instance of performance extrapolation occurs in neuroimaging studies, where the number of classes $k$ is
limited by experimental considerations.
Kay et al. [1] obtained fMRI brain scans which record how a single subject's visual cortex responds to natural images.
The label set $\mathcal{Z}$ corresponds to the space of all grayscale photographs of natural images,
and the set $\mathcal{Z}_1$ is a subset of 1750 photographs used in the experiment.
They construct a classifier which achieves over 0.75 accuracy for classifying the 1750 photographs;
based on exponential extrapolation, they estimate that it would take on the order of $10^{9.5}$ photographs
before the accuracy of the model drops below 0.10!  Directly validating this estimate would take immense resources,
so it would be useful to develop the theory needed to understand how to compute such extrapolations
in a principled way. 

However, in the fully general setting, it is impossible on construct
non-trivial bounds on the accuracy achieved on the new classes $\mathcal{Z}_K \setminus \mathcal{Z}_k$
based only on knowledge of $\mathcal{Z}_k$: after all, $\mathcal{Z}_k$ could consist entirely of well-separated classes
while the new classes $\mathcal{Z}_K \setminus \mathcal{Z}_k$ consist entirely of highly inseparable classes, or vice-versa.
Thus, the most important assumption for our theory is that of \emph{exchangeable sampling}.
The labels in $\mathcal{Z}_i$ are assumed to be an exchangeable sample from $\mathcal{Z}$.
The condition of exchangeability ensures that the separability of random subsets of $\mathcal{Z}$ can be inferred
by looking at the empirical distributions in $\mathcal{Z}_k$, and therefore that some estimate of the achievable
accuracy on $\mathcal{Z}_K$ can be obtained.

The assumption of exchangeability greatly limits the scope of application for our methods.
Many multi-class classification problems
have a hierarchical structure [2], or have class labels distributed according to
non-uniform discrete distributions, e.g. power laws [3]; in either case, exchangeability is violated.
It would be interesting to extend our theory to the hierarchical setting, or to handle non-hierarchical settings
with non-uniform prior class probabilities, but again we leave the subject for future work.

In addition to the assumption of exchangeability, we consider a restricted set of classifiers.
We focus on \emph{generative classifiers}, which are classifiers that work by training
a model separately on each class.  This convenient property 
allows us to characterize the accuracy of the classifier by selectively conditioning on one class at a time.
In section 3, we use this technique to reveal an equivalence between 
the expected accuracies of $\mathcal{Z}_k$ to moments of a common distribution.
This moment equivalence result allows standard approaches in statistics, such as U-statistics and
nonparametric pseudolikelilood, to be directly applied to the extrapolation problem, as we discuss in section 4.
In non-generative classifiers, the classification rule has a joint dependence on the entire set of classes,
and cannot be analyzed by conditioning on individual classes.
In section 5, we empirically study the performance of our classifiers.
Since generative classifiers only comprise a minority of the classifiers used in practice,
we applied our methods to a variety of generative and non-generative classifiers
in simulations and in one OCR dataset.  Our methods have varying success on generative and non-generative classifiers,
but seem to work badly for neural networks.

\noindent\emph{Our contribution.}

To our knowledge, we are the first to formalize the problem of prediction extrapolation.
We introduce three methods for prediction extrapolation: the method of extended unbiased estimation
and the constrained pseudolikelihood method are novel.  The third method, based
on asymptotics, is a new application of a recently proposed method 
for estimating mutual information [4].

\section{Setting}

Having motivated the problem of performance extrapolation,
we now reformulate the problem for notational and theoretical convenience.
Instead of requiring $\mathcal{Z}_k$ to be a random subset of $\mathcal{Z}$ as we did in section 1, take
$\mathcal{Z}=\mathbb{N}$ and $\mathcal{Z}_k = \{1,\hdots, k\}$.
We fix the size of $\mathcal{Z}_k$ without losing generality, since any monotonic sequence of 
finite subsets can be embedded in a sequence with $|\mathcal{Z}_k| = k$.
In addition, rather than randomizing the labels, we will randomize the marginal distribution $p(y|z)$ of each label;
Towards that end, let $\mathcal{Y} \subset \mathbb{R}^p$ be a space of feature vectors, and
let $\mathcal{P}(\mathcal{Y})$ be a measurable space of probability distributions on $\mathcal{Y}$.
Let $\mathbb{F}$ be a probability measure on $\mathcal{P}$,
and let $F_1, F_2,\hdots$ be an infinite sequence of i.i.d. draws from $\mathbb{F}$.
We refer to $\mathbb{F}$, a probability measure on probability measures, as a \emph{meta-distribution}.
The distributions $F_1,\hdots, F_k$ are the marginal distributions of the first $k$ classes.
Further assuming that the labels are equiprobable, we rewrite the accuracy as
\[
\text{acc}^{(t)} = \frac{1}{t}\sum_{i=1}^t \Pr_{F_i}[h^{(t)}(Y) = i].
\]
where the probabilities are taken over $Y \sim F_i$.

In order to construct the classification rule $h^{(t)}$, we need data from the classes $F_1,\hdots, F_t$.
In most instances of multi-class classification, one observes independent observations from each $F_i$
which are used to construct the classifier.  Since the order of the observations
does not generally matter, a sufficient statistic for the training data for the $t$-th classification problem
is the collection of empirical distributions
$\hat{F}_1^{(t)},\hdots,\hat{F}_t^{(t)}$ for each class.
Henceforth, we make the simplifying assumption that the training data for the $i$-th class remains fixed
from $t =i, i+1,\hdots$, so we drop the superscript on $\hat{F}_i^{(t)}$.
Write $\hat{\mathbb{F}}(F)$ for the conditional distribution of $\hat{F}_i$ given  $F_i = F$;
also write $\hat{\mathbb{F}}$ for the marginal distribution of $\hat{F}$ when $F \sim \mathbb{F}.$
As an example, suppose every class has the number of training examples $r \in \mathbb{N}$; then $\hat{F}$
is the empirical distribution of $r$ i.i.d. observations from $F$, and $\hat{\mathbb{F}}(F)$ is the \emph{empirical meta-distribution} of $\hat{F}$.
Meanwhile, $\hat{\mathbb{F}}$ is the true meta-distribution of the empirical distribution of $r$ i.i.d. draws from a random $F \sim \mathbb{F}$.

\subsection{Multiclass classification}

Extending the formalism of Tewari and Bartlett [5]\footnote{As in their framework,
we define a classifier as a vector-valued function.  However, we introduce the notion of a classifier as a multiple-argument functional on empirical distributions, which echoes the functional formulation of estimators common in the statistical literature.},
we define a classifier as a collection of mappings
$\mathcal{M}_i: \mathcal{P}(\mathcal{Y})^k \times \mathcal{Y} \to \mathbb{R}$ called \emph{classification functions.}
Intuitively speaking, each classification function \emph{learns a model} from the first $k$ arguments, which are
the empirical marginals of the $k$ classes, $\hat{F}_1,\hdots, \hat{F}_k$.  For each class, the classifier assigns a real-valued \emph{classification score} to the \emph{query point} $y \in \mathcal{Y}$.  A higher score $\mathcal{M}_i(\hat{F}_1,\hdots, \hat{F}_k, y)$ indicates a higher estimated probability that $y$ belongs to the $k$-th class.  
Therefore, the classification rule corresponding to a classifier $\mathcal{M}_i$ assigns
a class with maximum classification score to $y$:
\[
h(y) = \argmax_{i \in \{1,\hdots, k\}} \mathcal{M}_i(y).
\]
For some classifiers, the classification functions $\mathcal{M}_i$ are especially simple
in that $\mathcal{M}_i$ is only a function of $\hat{F}_i$ and $y$.
Furthermore, due to symmetry, in such cases one can write
\[
\mathcal{M}_i(\hat{F}_1,\hdots, \hat{F}_k, y) = \mathcal{Q}(\hat{F}_i, y),
\]
where $\mathcal{Q}$ is called a \emph{single-class classification function} (or simply \emph{classification function}),
and we say that $\mathcal{M}$ is a \emph{generative classifier}.
Quadratic discriminant analysis and Naive Bayes [6] are two examples of
generative classifiers\footnote{For QDA, the classification function is given by
\[
\mathcal{Q}_{QDA}(\hat{F}, y) = -(y - \mu(\hat{F}))^T \Sigma(\hat{F})^{-1} (y-\mu(\hat{F})) - \log\det(\Sigma(\hat{F})),
\]
where $\mu(F) = \int y dF(y)$ and $\Sigma(F) = \int (y-\mu(F))(y-\mu(F))^T dF(y)$.
In Naive Bayes, the classification function is
\[
\mathcal{Q}_{NB}(\hat{F},  y) = \sum_{i=1}^n \log \hat{f}_i(y_i),
\]
where $\hat{f}_i$ is a density estimate for the $i$-th component of
$\hat{F}$.}.
The \emph{generative} property allows us to prove strong results about the accuracy of the classifier
under the exchangeable sampling assumption, as we see in Section 3.

\section{Performance extrapolation for generative classifiers}

Let us specialize to the case of a generative classifier, with classification function $\mathcal{Q}$.
Consider estimating the expected accuracy for the $k$-th classification problem,
\begin{equation}\label{eq:pt}
p_k \stackrel{def}{=} \E[\text{acc}^{(k)}].\end{equation}
In the case of a generative classifier, we have
\[
p_k = \E[acc^{(k)}] = \E\left[\frac{1}{k}\sum_{i=1}^k \Pr_{Y \sim F_i}[\mathcal{Q}(\hat{F}_i, Y) > \max_{j \neq i}\mathcal{Q}(\hat{F}_j, Y)]\right].
\]
Define the \emph{conditional accuracy} function $u(\hat{F}, y)$ which maps a
distribution $\hat{F}$ on $\mathcal{Y}$ and a \emph{test} observation $y$ to
a real number in $[0,1]$.  The conditional accuracy gives the
probability that for independently drawn $\hat{F}'$ from $\hat{\mathbb{F}}$, that
$\mathcal{Q}(\hat{F}, y)$ will be greater than $\mathcal{Q}(\hat{F}', y)$:
\[
u(\hat{F}, y) = \Pr_{\hat{F}' \sim \hat{\mathbb{F}}}[\mathcal{Q}(\hat{F}, y) > \mathcal{Q}(\hat{F}', y)].
\]
Define the \emph{conditional accuracy} distribution $\nu$ as the law
of $u(\hat{F}, Y)$ where $\hat{F}$ and $Y$ are generated as follows:
(i) a true distribution $F$ is drawn from $\mathbb{F}$; 
(ii) the empirical distribution $\hat{F}$ is drawn from $\hat{\mathbb{F}}(F)$ (i.e., the training data for the class),
(iii) the query $Y$ is drawn from $F$, with $Y$ independent of $\hat{F}$ (i.e. a single test data point from the same class.)
The significance of the conditional accuracy
distribution is that the expected accuracy $p_t$ can be
written in terms of its moments.

\noindent\textbf{Theorem 3.1.} \emph{
Let $\mathcal{Q}$ be a single-distribution classification function, and let $\mathbb{F}$, $\hat{\mathbb{F}}(F)$ be a distribution on $\mathcal{P}(\mathcal{Y}).$
Further assume that
$\hat{\mathbb{F}}$ and $\mathcal{Q}$ jointly satisfy the
\emph{tie-breaking} property:
\begin{equation}\label{eq:tie}
\Pr[\mathcal{Q}(\hat{F}, y) = \mathcal{Q}(\hat{F}', y)] = 0
\end{equation}
for all $y \in \mathcal{Y}$, where $\hat{F}, \hat{F}' \stackrel{iid}{\sim} \hat{\mathbb{F}}$.
Let $U$ be defined as the random variable
$U = u(\hat{F}, Y)$
for $F \sim \mathbb{F}$, $Y \sim F$, and $\hat{F} \sim \hat{\mathbb{F}}(F)$ with $Y \perp \hat{F}$.
Then \[p_k = \E[U^{k-1}],\]
where $p_k$ is the expected accuracy as defined by \eqref{eq:pt}.
}

\noindent\textbf{Proof.}  
Write $q^{(i)}(y) = \mathcal{Q}(\hat{F}_i, y)$.
By using conditioning and
conditional independence, $p_k$ can be written
\begin{align*}
p_k &= \E\left[ \frac{1}{k}\sum_{i=1}^k  \Pr_{F_i}[q^{(i)}(Y) > \max_{j\neq i} q^{(j)}(Y)] \right]
\\&= \E\left[ \Pr_{F_1}[q^{(1)}(Y) > \max_{j\neq 1} q^{(j)}(Y)] \right]
\\&= \E_{F_1}[\Pr[q^{(1)}(Y) > \max_{j\neq 1} q^{(j)}(Y)|\hat{F}_1, Y]]
\\&= \E_{F_1}[\Pr[\cap_{j > 1} q^{(1)}(Y) > q^{(j)}(Y)|\hat{F}_1, Y]]
\\&= \E_{F_1}[\prod_{j > 1}\Pr[q^{(1)}(Y) > q^{(j)}(Y)|\hat{F}_1, Y]]
\\&= \E_{F_1}[\Pr[q^{(1)}(Y) > q^{(2)}(Y)|\hat{F}_1, Y]^{k-1}]
\\&= \E_{F_1}[u(\hat{F}_1, Y)^{k-1}] = \E[U^{k-1}].
\end{align*}
$\Box$

Theorem 3.1 tells us that the problem of extrapolation can be
approached by attempting to estimate the conditional accuracy
distribution.  The $(t-1)$-th moment of $U$ gives us $p_t$, which will
in turn be a good estimate of $\text{acc}^{(t)}$.

While $U = u(\hat{F}, Y)$ is not directly observed, we can obtain unbiased estimates of $u(\hat{F}_i, y)$
by using test data.  For any $\hat{F}_1,\hdots, \hat{F}_k$, and independent test point $Y \sim F_i$, define
\begin{equation}\label{eq:hatu}
\hat{u}(\hat{F}_i, Y) = \frac{1}{k -1}\sum_{j \neq i} I(\mathcal{Q}(\hat{F}_i, Y) > \mathcal{Q}(\hat{F}_j, Y)).
\end{equation}
Then $\hat{u}(\hat{F}_i, Y)$ is an unbiased estimate of $u(\hat{F}_i, Y)$, as stated in the following theorem.

\noindent\textbf{Theorem 3.2.}\emph{
Assume the conditions of theorem 3.1.
Then defining 
\begin{equation}\label{eq:veq}
V = (k-1)\hat{u}(\hat{F}_i, y),\end{equation}
we have
\[V \sim \text{\emph{Binomial}}(k-1, u(\hat{F}_i, y)).\]
Hence,
\[\E[\hat{u}(\hat{F}_i, y)] = u(\hat{F}_i, y).\]
}

In section 4, we will use this result to estimate the moments of $U$.
Meanwhile, since $U$ is a random variable on $[0, 1]$, we also conclude that $p_t$ follows a \emph{mixed exponential decay}.
Let $\alpha$ be the law of $-\log(U)$.
Then from change-of-variables $\kappa =-\log(u)$, we get
\[p_t = \E[U^{t-1}] = 
\int_0^1 u^{t-1} d\nu(u) = \int_0^1 e^{t\log(u)} \frac{1}{u}d\nu(u) = 
\int_{\mathbb{R}^{+}} e^{-\kappa t} d\alpha(\kappa).\]
This fact immediately suggests the technique of fitting a mixture of exponentials to the test accuracy at $t =2,3,\hdots, k$:
we explore this idea further in Section 4.1.

\subsection{Properties of the conditional accuracy distribution}

The conditional accuracy distribution $\nu$ is determined by $\mathbb{F}$
and $\mathcal{Q}$.  What can we say about the the conditional accuracy
distribution without making any assumptions on either $\mathbb{F}$ or
$\mathcal{Q}$?  The answer is: not much.  For an arbitrary probability
measure $\nu'$ on $[0,1]$, one can construct $\mathbb{F}$ and
$\mathcal{Q}$ such that the conditional accuracy $U$ has the distribution $\nu'$, even if one makes the \emph{perfect sampling assumption} that $\hat{F}=F.$

\noindent\textbf{Theorem 3.3.} \emph{ Let $U$ be defined as in Theorem
  3.1, and let $\nu$ denote the law of $U$.  Then, for any probability
  distribution $\nu'$ on $[0,1]$, one can construct a
  meta-distribution $\mathbb{F}$ and a classification function $\mathcal{Q}$ such
  that the conditional accuracy $U$ has distribution $\nu'$ under perfect sampling (that is, $\hat{F} = F$.)  }

\textbf{Proof.}  
Let $G$ be the cdf of $\nu$, $G(x) = \int_0^x d\nu(x)$, and let $H(u) = \sup_x \{G(x) \leq u\}$.
Define $\mathcal{Q}$ by
\[
\mathcal{Q}(\hat{F}, y) = \begin{cases}
0 &\text{ if }\mu(\hat{F}) > y + H(y)\\
0 & \text{ if }y + H(y) > 1 \text{ and }\mu(\hat{F}) \in [H(y) - y, y]\\
1 + \mu(\hat{F}) - y &\text{ if } \mu(\hat{F}) \in [y, y + H(y)]\\
1 + y + \mu(\hat{F}) &\text{ if }\mu(\hat{F}) + H(y) > 1 \text{ and }\mu(\hat{F}) \in [0, H(y) - y]. 
\end{cases}
\]
Let $\theta \sim \text{Uniform}[0,1]$,
and define $F \sim \mathbb{F}$ by $F = \delta_\theta$, and also $\hat{F} = F.$
A straightforward calculation yields that $\nu = \nu'$. $\Box$

On the other hand, we can obtain a positive result if we assume that
the classifier approximates a \emph{Bayes classifier.}
Assuming that $F$ is absolutely continuous with respect to Lebesgue measure $\Lambda$ with probability one,
a Bayes classifier results from assuming perfect sampling ($\hat{F} = F$) and taking
$\mathcal{Q}(\hat{F}, y) = \frac{dF}{d\Lambda}(y)$.
Theorem 3.4. states that for a Bayes classifier, the measure $\nu$ has a density $\eta(u)$ which is monotonically increasing.
Since a `good' classifier approximates the Bayes classifier, we intuitively expect that a monotonically
increasing density $\eta$ is a good model for the conditional accuracy distribution of a `good' classifier.

\noindent\textbf{Theorem 3.4.} \emph{ Assume the conditions of theorem 3.1, and further suppose
that $\hat{F} = F$, $F$ is absolutely continuous with respect to $\Lambda$ with probability one,
that $\mathcal{Q}(\hat{F}, y) = \frac{dF}{d\Lambda}(y)$, and that $F|Y$ has a regular conditional probability distribution.
Let $\nu$ denote the law of $U$.    Then $\nu$ has a density $\eta(u)$ on $[0, 1]$ which is monotonic in $u$.
}

\noindent\textbf{Proof.}
It suffices to prove that
\[
\nu([u, u + \delta]) < \nu([v, v + \delta])
\]
for all $0 < u < v < 1$ and $0 < \delta < 1-v$.
Let $\mathcal{P}_{ac}(\mathcal{Y})$ denote the space of distributions supported on $\mathcal{Y}$ which are
absolutely continuous with respect to $p$-dimensional Lebesgue measure $\Lambda$.
Let $\mathbb{Y}$ denote the marginal distribution of $Y$ for $Y \sim F$ with $F \sim \mathbb{F}$.
Define the set 
\[
J_y(A) =\{F \in \mathcal{P}_{ac}(\mathcal{Y}): u(F, y) \in A\}.
\]
for all $A \subset [0, 1].$
One can verify that for all $y \in \mathcal{Y}$,
\[
\Pr_\mathbb{F}[J_y([u, u + \delta])|Y=y] \leq \Pr_\mathbb{F}[J_y([v, v + \delta])|Y=y],
\]
using the fact that $\mathbb{F}$ has no atoms.  Hence, we obtain
\[
\Pr[U \in [u-\delta, u + \delta]] = \E_{\mathbb{Y}}[\Pr_\mathbb{F}[J_Y([u, u + \delta])|Y]] 
\leq \E_{\mathbb{Y}}[\Pr_\mathbb{F}[J_Y([v, v + \delta])|Y]]  = Pr[U \in [v - \delta, v + \delta]].
\]
Taking $\delta \to 0$, we conclude the theorem. $\Box$\newline

\section{Estimation}

Suppose we have $m$ independent test repeats per class, $y^{(i),1}\hdots, y^{(i), m}$.
Let us define
\[
V_{i,j} = \sum_{\ell\neq i} I(\mathcal{M}_i(\hat{F}_1,\hdots, \hat{F}_k, y^{(i, j)})  > \mathcal{M}_\ell(\hat{F}_1,\hdots, \hat{F}_k, y^{(i, j)})),
\]
which coincides with the definition \eqref{eq:veq} in the special case that $\mathcal{M}$ is generative.

At a high level, we have a hierarchical model where $U$ is drawn from a distribution $\nu$ on $[0, 1]$
and then $V_{i, j} \sim \text{Binomial}(k, U)$.
Let us assume that $U$ has a density $\eta(u)$: then the marginal distribution of $V_{i, j}$ can be written
\[
\Pr[V_{i,j} = \ell] = \begin{pmatrix}
k \\ \ell
\end{pmatrix}
\int_0^1 u^\ell (1-u)^{k-\ell} \eta(u) du.
\]
However, the observed $\{V_{i, j}\}$ do \emph{not} comprise an i.i.d. sample.

We discuss the following three approaches for estimating $p_t =
\E[U^{t-1}]$ based on $V_{i, j}$.  The first is an extension of \emph{unbiased
  estimation} based on binomial U-statistics, which is discussed in
Section 4.1.  The second is the \emph{pseudolikelihood} approach.  In
problems where the marginal distributions are known, but the
dependence structure between variables is unknown, the
\emph{pseudolikelihood} is defined as the product of the marginal
distributions.  For certain problems in time series analysis and
spatial statistics, the maximum pseudolikelihood estimator (MPLE) is
proved to be consistent [7].  We discuss pseudolikelihood-based
approaches in Section 4.2.  Thirdly, we note that the high-dimensional
theory of Anon 2016 [4] can be applied for prediction accuracy, which we discuss in Section 4.3.

\subsection{Extensions of unbiased estimation}

If $V \sim \text{Binomial}(k, U)$, then an unbiased estimator of $U^t$ exists
if and only if $0 \leq t \leq k$.

The theory of U-statistics [8] provides the minimal variance unbiased estimator for $U^t$:
\[
U^t = \E\left[\begin{pmatrix}
V \\ t
\end{pmatrix}
\begin{pmatrix}
k \\ t
\end{pmatrix}^{-1}\right].
\]

This result can be immediately applied to yield an unbiased estimator of $p_t$, when $t \leq k$:
\begin{equation}\label{eq:ustat}
\hat{p}_t^{UN} =  \frac{1}{km}\sum_{i=1}^k\sum_{j=1}^{m} \begin{pmatrix}
V_{i, j} \\ t-1
\end{pmatrix}
\begin{pmatrix}
k \\ t-1
\end{pmatrix}^{-1}.
\end{equation}
However, since $\hat{p}_t^{UN}$ is undefined for $k \geq t$, we can use exponential extrapolation
to define an extended estimator $\hat{p}_t^{EXP}$ for $k > t$.
Let $\hat{\alpha}$ be a measure defined by solving the optimization problem
\[
\text{minimize}_{\alpha} \sum_{t=2}^{k} \left(\hat{p}_t^{UN} - \int_0^\infty \exp[-t\kappa] d\alpha(\kappa)\right)^2.
\]
After discretizing the measure $\hat{\alpha}$, we obtain a convex optimization problem
which can be solved using non-negative least squares [9].
Then define
\[
\hat{p}_t^{EXP} = \begin{cases}
\hat{p}_t^{UN}&\text{ for }t \leq k,\\
\int_0^\infty \exp[-t\kappa] d\hat{\alpha}(\kappa))&\text{ for }t > k.
\end{cases}
\]

\subsection{Maximum pseudolikelihood}

\begin{figure}
\centering
\begin{tabular}{ccrl}
Estimated density $\hat{\eta}$ &Estimated moment $\E[U^t]$ & \\
\multirow{5}{*}{\includegraphics[scale = 0.5, clip=true, trim=0.2in 0.6in 0 0.7in]{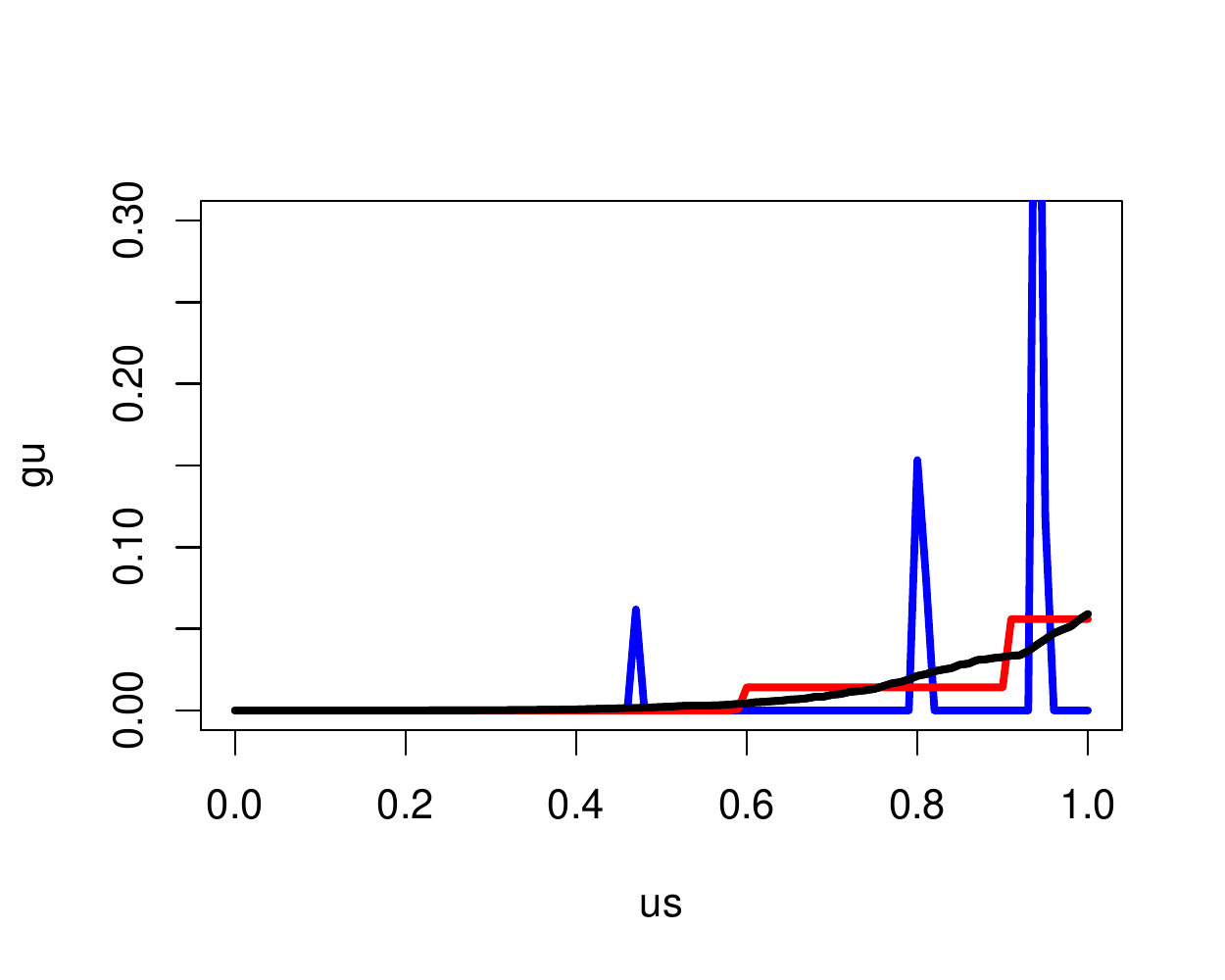}} &
\multirow{5}{*}{\includegraphics[scale = 0.5, clip=true, trim=0.2in 0.6in 0 0.7in]{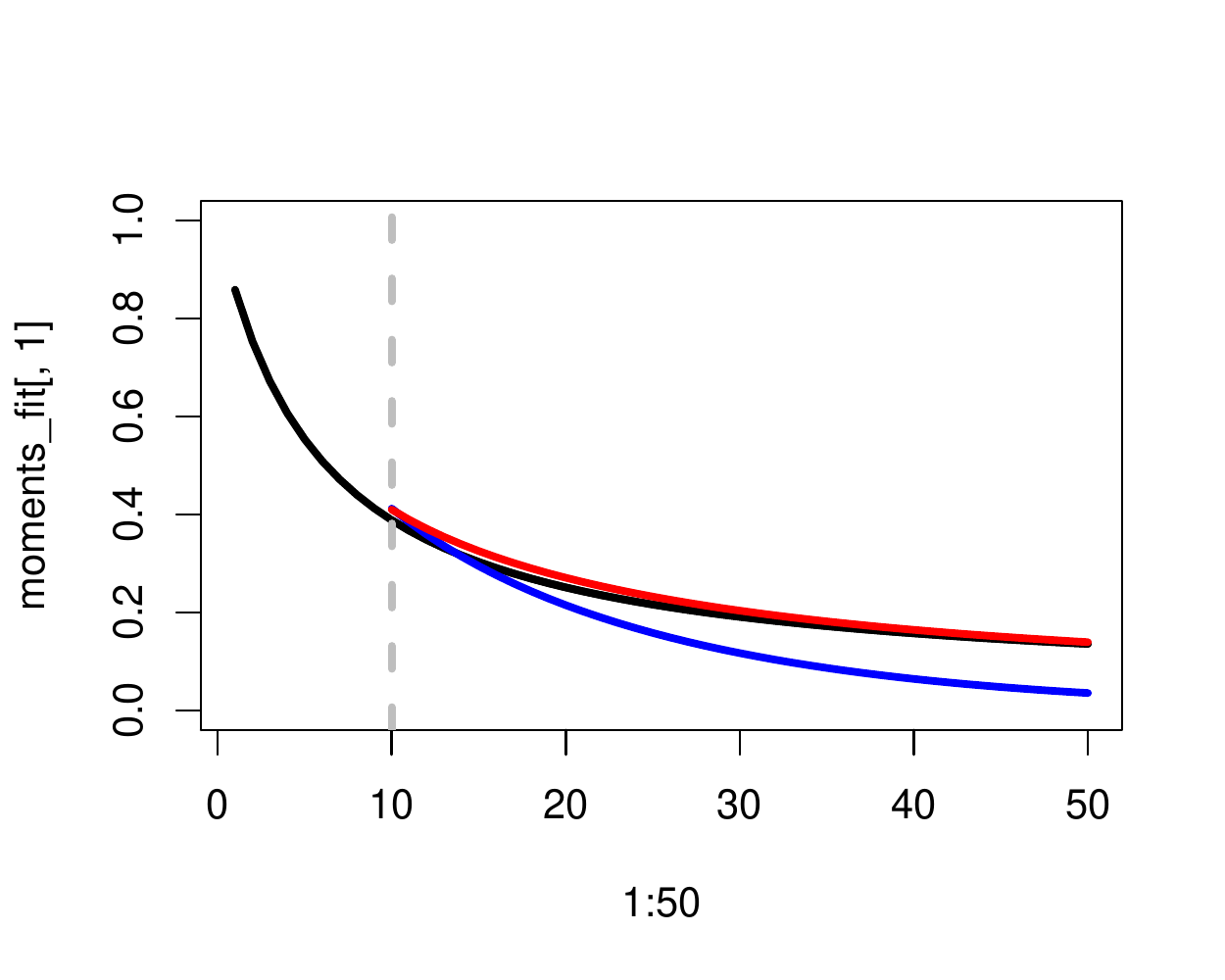}} & & \\
& & \crule[black]{0.2cm}{0.2cm} & Truth\\
& & & \\
& & \crule[blue]{0.2cm}{0.2cm} & MPLE \\
& & & \\
& & \crule[red]{0.2cm}{0.2cm} & CONS \\
& & & \\
& & & \\
& & & \\
$u$& $t$& & \\
\end{tabular}
\caption{Maximum pseudolikelihood (MPLE) versus constrained pseudolikelihood (CONS).
Adding constraints improves the estimation of the density $\eta(u)$, as well as moment estimation.}
\end{figure}

The (log) pseudolikelihood is defined as
\begin{equation}\label{eq:psuedo}
\ell(\eta) = \sum_{i=1}^k \sum_{j=1}^{m} \log\left(\int u^{V_{i, j}} (1-u)^{k - V_{i, j}} \eta(u) du\right),
\end{equation}
and a maximum pseudolikelihood estimator (MPLE) is defined as any
density $\hat{\eta}$ such that
\[
\ell(\hat{\eta}_{MPLE}) = \sup_{\eta} \ell(\eta).
\]
The motivation for $\hat{\eta}_{MPLE}$ is that it consistently
estimates $\eta$ in the limit where $k \to \infty$.
However, in finite samples, $\hat{\eta}_{MPLE}$ is not uniquely defined,
and if we define the plug-in estimator
\[
\hat{p}_t^{MPLE} = \int u^{t-1} \hat{\eta}_{MPLE}(u) du,
\]
$\hat{p}_t^{MPLE}$ can vary over a large range, depending on which $\hat{\eta} \in \argmax_{\eta} \ell_t(\eta)$
is selected.
These shortcomings motivate the adoption of additional constraints on the estimator $\hat{\eta}$.

Theorem 3.4. motivates the \emph{monotonicity constraint} that $\frac{d\hat{\eta}}{du} > 0$.
A second constraint is to restrict the $k$-th moment of $\hat{\eta}$ to match the unbiased estimate.
The addition of these constraints yields the constrained PMLE
$\hat{\eta}_{CON}$, which is obtained by solving
\[
\text{maximize }\ell(\eta) \text{ subject to }\int u^{k-1} \eta(u) du = \hat{p}_k^{UN}\text{ and }\frac{d\hat{\eta}}{du} > 0.
\]
By discretizing $\eta$, all of the above maximization problems can be solved using a general-purpose convex solver\footnote{
We found that the disciplined convex programming language CVX, using the ECOS second-order cone programming solver,
succeeds in optimizing the problems where the dimension of the discretized $\eta$ is as large as 10,000 [10, 11].}.
While the added constraints do not guarantee a unique solution,
they improve estimation of $\eta$ and thus improve moment estimation (Figure 1.)

\subsection{High-dimensional asymptotics}

Under a number of conditions on the distribution $\mathbb{F}$, including (but not limited to) having a large dimension $p$,
Anon [4] relate the accuracy $p_t$ of the Bayes classifier to the mutual information between the label $z$ and
the response $y$:
\[
p_t = \bar{\pi}_t(\sqrt{2I(Z; Y)}).
\]
where
\[
\bar{\pi}_k(c) = \int_{\mathbb{R}} \phi(z - c)  \Phi(z)^{k-1} dz.
\]
While our goal is not to estimate the mutual information, we note that the results of Anon 2016
imply a relationship between $p_k$ and $p_K$ for the Bayes accuracy under the high-dimensional regime:
\[
p_K = \bar{\pi}_K\left(\bar{\pi}_k^{-1}(p_k)\right).
\]
Therefore, under the high-dimensional conditions of [4] and assuming that the classifier approximates
the Bayes classifier, we naturally obtain the following estimator
\[
\hat{p}_t^{HD} = \bar{\pi}_K\left(\bar{\pi}_k^{-1}(\hat{p}_k^{UN})\right).
\]

\section{Results}

We applied the methods described in Section 4 on a simulated gaussian mixture (Figure 2)
and on a Telugu character classification task [12] (Table 1.)

For the simulated gaussian mixture, we vary the size of the initial subset from $k=3$ classes to $k=K=50$ classes,
and extrapolate the performance for gaussian mixture model, multinomial logistic, and one-layer neural network (with 10 sigmoidal units.)
Figure 3 shows how the predicted $K$-class accuracy changes as $k$ is varied.
We see that the predicted accuracy curves for QDA and Logistic have similar behavior,
even though QDA is generative and multinomial logistic is not.  All three methods perform better on QDA and logistic classifiers
than on the neural network: in fact, for the neural network, the test accuracy of the initial set, $\text{acc}^{(k)}$,
becomes a better estimator of $\text{acc}^{(K)}$ than the three proposed methods for most of the curve.
We also see that the exponential extrapolation method, $\hat{p}^{EXP}$,
is more variable than constrained pseudolikelihood $\hat{p}^{CONS}$ and high-dimensional estimator $\hat{p}^{HD}$.
Additional simulation results can be found in the supplement.

In the character classification task, we predict the 400-class accuracy of naive Bayes, multinomial
logistic regression, SVM [6], $\epsilon$-nearest neighbors\footnote{$k$-nearest neighbors with $k = \epsilon n$ for fixed $\epsilon > 0$}, and deep neural networks\footnote{The network architecture is as follows: 
{\tt 48x48-4C3-MP2-6C3-8C3-MP2-32C3-50C3-MP2-200C3-SM.}
48x48 binary input image, $m$C3 is a 3x3 convolutional layer with $m$ output maps, MP2 is a 2x2 max-pooling layer, and SM is a softmax output layer on 20 or 400 classes.} using 20-class data with 103 training examples per class (Table 1).
Taking the test accuracy on 400 classes (using 50 test examples per class) as a proxy for $\text{acc}^{(400)}$,
we compare the performance of the three extrapolation methods; as a benchmark,
also consider using the test accuracy on 20 classes as an estimate.
The exponential extrapolation method performs well only for the deep neural network.  Meanwhile, constrained PMLE achieves accurate extrapolation for two out of four classifiers: logistic and SVM
but failed to converge for the the deep neural network (due to the high test accuracy).
The high-dimensional estimator $\hat{p}^{HD}$  performs well on the multinomial logistic, SVM, and deep neural network classifiers.  All three methods beat the benchmark (taking the test accuracy at 20) for the first four classifiers;
however, the benchmark is the best estimator for the deep neural network,
similarly to what we observe in the simulation (albeit with a shallow network rather than a deep network.)

\begin{figure}
\centering
\begin{tabular}{cccrl}
QDA & Logistic & Neural Net & \\
\multirow{5}{*}{\includegraphics[scale = 0.5, clip=true, trim=0.2in 0.6in 0.2in 0.7in]{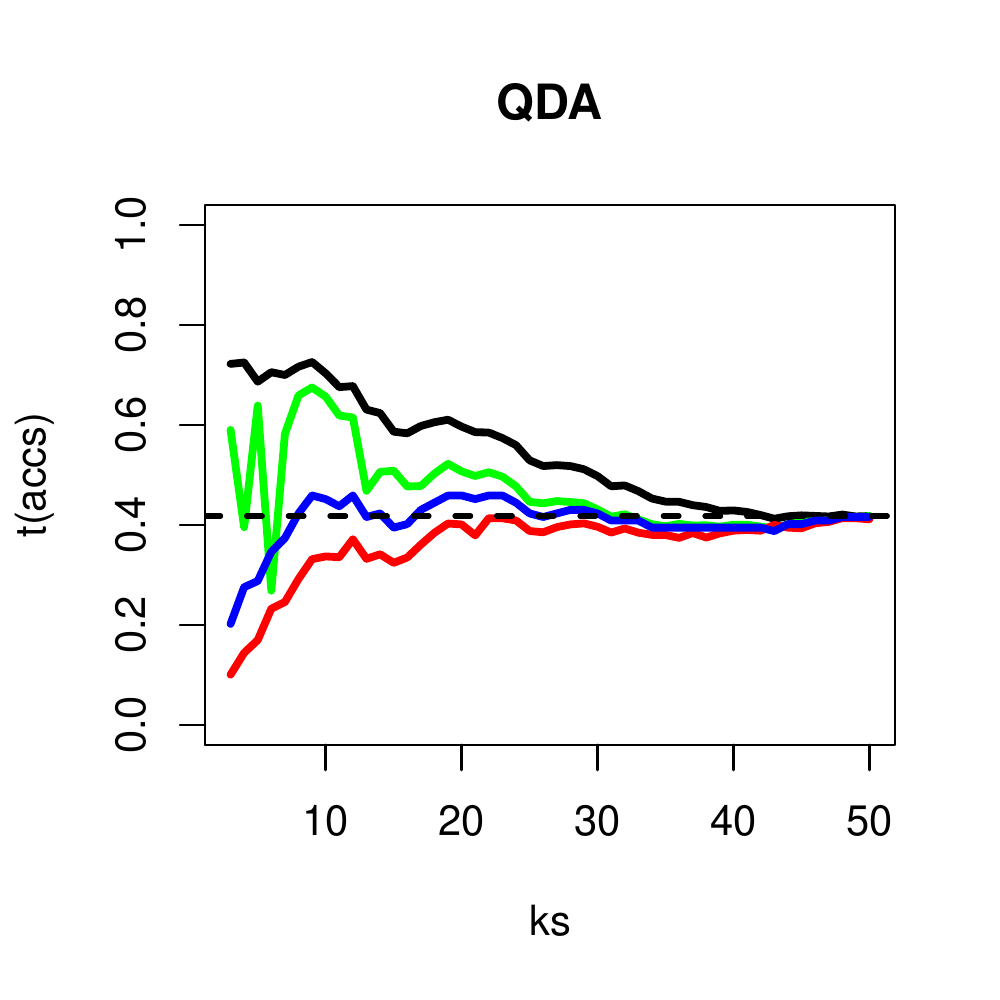}} &
\multirow{5}{*}{\includegraphics[scale = 0.5, clip=true, trim=0.75in 0.6in 0.2in 0.7in]{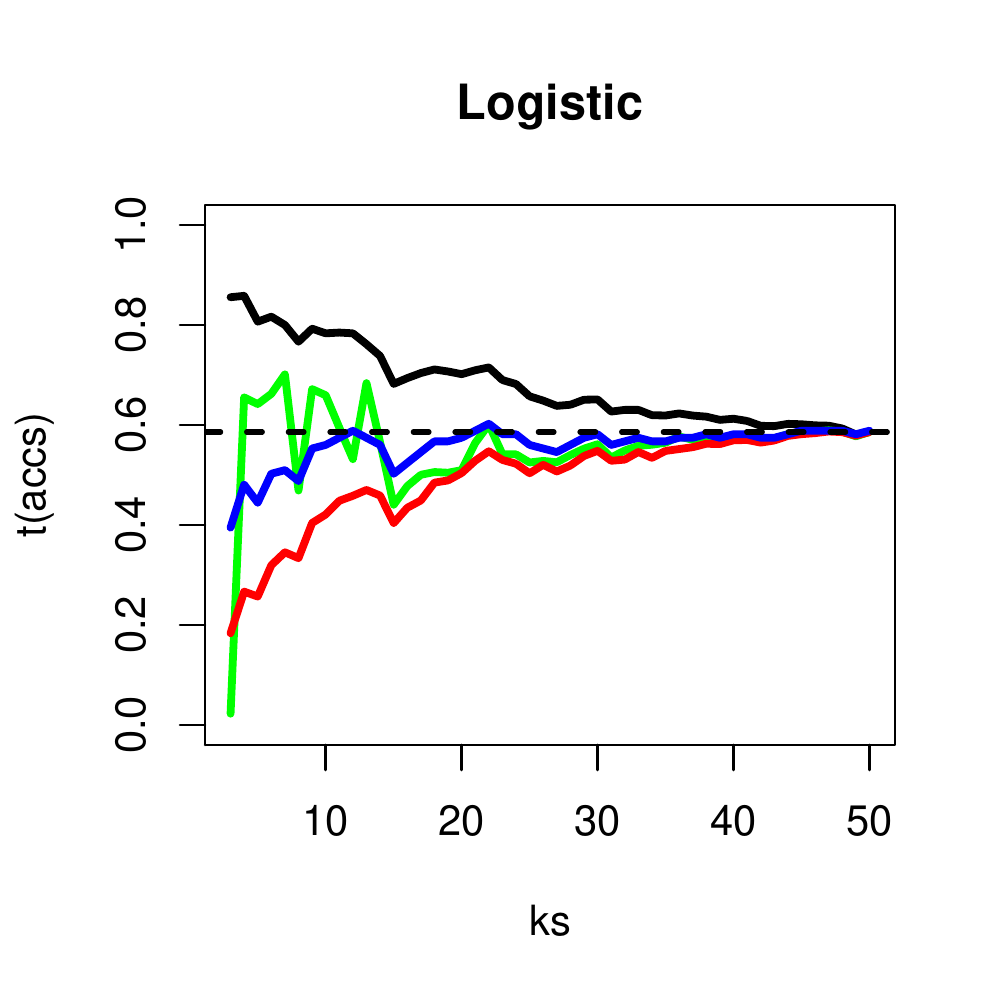}} & 
\multirow{5}{*}{\includegraphics[scale = 0.5, clip=true, trim=0.75in 0.6in 0.2in 0.7in]{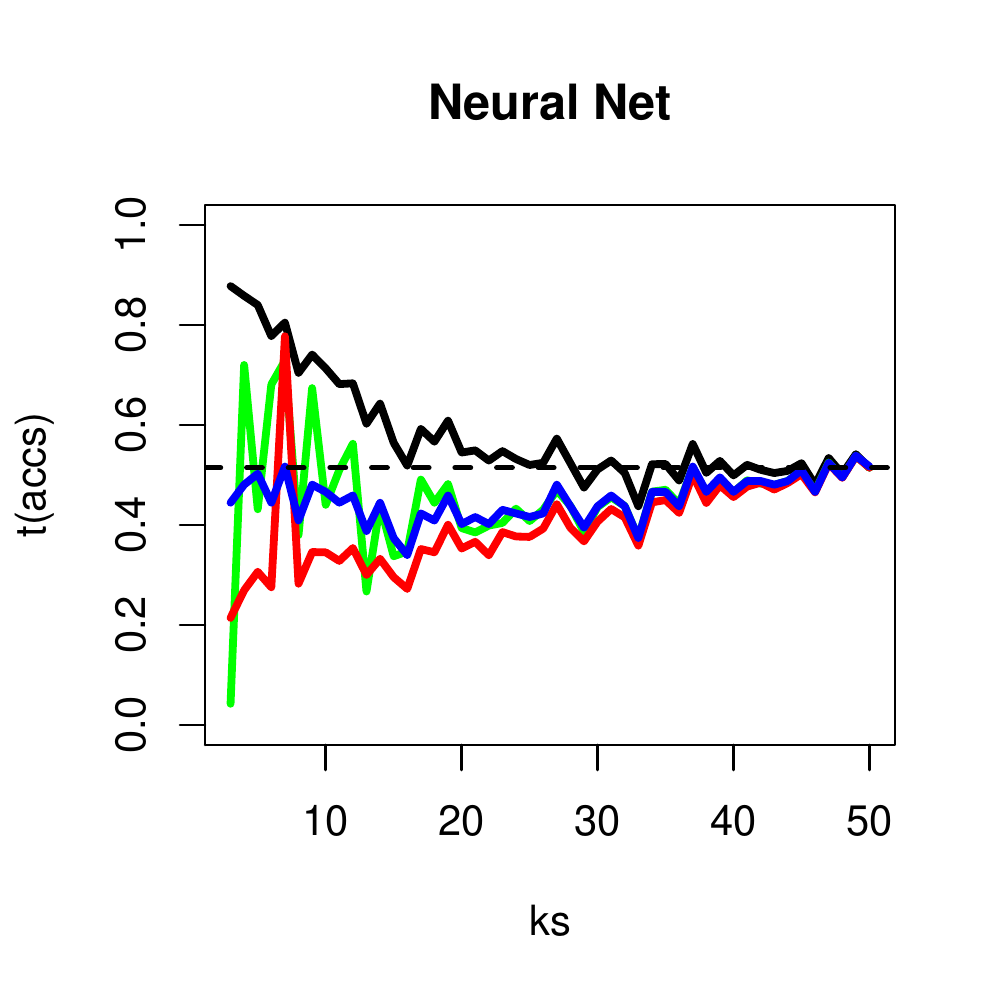}} & 
& \\
&& & &\\
&& &  \crule[black]{0.2cm}{0.2cm} & $\text{acc}^{(k)}$ \\
&& & \crule[green]{0.2cm}{0.2cm} & $\hat{p}^{EXP}$ \\
&& & \crule[red]{0.2cm}{0.2cm} & $\hat{p}^{CONS}$ \\
&& & \crule[blue]{0.2cm}{0.2cm} & $\hat{p}^{HD}$ \\
&& & & \\
&& & & \\
&& & & \\
$k$ &$k$& $k$& & \\
\end{tabular}
\caption{Predictions for $\text{acc}^{(50)}$ as $k$, the size of the subset, is varied.  Our methods work better for QDA and Logistic than Neural Net; overall, $\hat{p}^{EXP}$ has higher variability than $\hat{p}^{CONS}$ and $\hat{p}^{HD}$.}
\end{figure}

\begin{table}
\centering
\begin{tabular}{|c||c|c|c|c|c|}\hline
Classifier & Test $\text{acc}^{(20)}$ & Test $\text{acc}^{(400)}$ & $\hat{p}^{EXP}_{400}$ & $\hat{p}^{CONS}_{400}$ & $\hat{p}^{HD}_{400}$\\ \hline
Naive Bayes & 0.947 & 0.601 & 0.884 & \textbf{0.659} & 0.769 \\ \hline
Logistic & 0.922 & 0.711 & 0.844 & \textbf{0.682} & 0.686 \\ \hline
SVM & 0.860 & 0.545 & 0.737 & 0.473 & \textbf{0.546} \\ \hline
$\epsilon$-NN & 0.964 & 0.591 & 0.895 & \textbf{0.395} & 0.839\\ \hline
Deep neural net & \textbf{0.995} & 0.986 & 0.973 & (*) & 0.957\\ \hline
\end{tabular}
\caption{Performance extrapolation: predicting the accuracy on 400 classes using data from 20 classes on a Telugu character dataset. (*) indicates failure to converge.
$\epsilon = 0.002$ for $\epsilon$-nearest neighbors.}
\end{table}

\section{Discussion}

Empirical results indicate that our methods generalize beyond generative classifiers.
A possible explanation is that since the Bayes classifier is generative,
any classifier which approximates the Bayes classifier is also `approximately generative.'
However, an important caveat is that the classifier must already attain close to the Bayes accuracy
on the smaller subset of classes.  If the classifier is initially far from the Bayes classifier,
and then becomes more accurate as more classes are added, our theory could underestimate the
accuracy on the larger subset.  This is a non-issue for generative classifiers when the training data per class is fixed,
since a generative classifier approximates the Bayes rule if and only if the single-class classification function approximates the
Bayes optimal single-class classification function.  On the other hand, for classifiers with built-in \emph{model selection}
or \emph{representation learning}, it is expected that the classification functions become more accurate,
in the sense that they better approximate a monotonic function of the Bayes classification functions,
as data from more classes is added.

Our results are still too inconclusive for us to recommend the use of any of these estimators in practice.
Theoretically, it still remains to derive confidence bounds for the generative case;
practically, additional experiments are needed to establish the reliability of these estimators
in specific applications.  There also remains plenty of room for new and improved estimators in this area:
for instance, fixing the instability of the constrained pseudolikelihood estimator when the test accuracy is high.


\subsubsection*{Acknowledgments}

We thank John Duchi, Steve Mussmann, Qingyun Sun, Jonathan Taylor, Trevor Hastie, Robert Tibshirani for useful discussion.  CZ is supported by an NSF graduate research fellowship.
\newpage
\section*{References}

\small

[1] Kay, K. N., Naselaris, T., Prenger, R. J., \& Gallant, J. L. (2008). ``Identifying natural images from human brain activity.'' 
\emph{Nature}, 452(March), 352-355.

[2] Deng, J., Berg, A. C., Li, K., \& Fei-Fei, L. (2010). ``What does classifying more than 10,000 image categories tell us?'' \emph{Lecture Notes in Computer Science}, 6315 LNCS(PART 5), 71-84. 

[3] Garfield, S., Stefan W., \& Devlin, S. (2005). ``Spoken language classification using hybrid classifier combination." 
\emph{International Journal of Hybrid Intelligent Systems} 2.1: 13-33.

[4] Anonymous, A. (2016). ``Estimating mutual information in high dimensions via classification error.''  Submitted to 
\emph{NIPS 2016.}

[5] Tewari, A., \& Bartlett, P. L. (2007). ``On the Consistency of Multiclass Classification Methods.''
\emph{Journal of Machine Learning Research}, 8, 1007-1025.

[6] Hastie, T., Tibshirani, R., \& Friedman, J., (2008). \emph{The elements
of statistical learning.} Vol. 1. Springer, Berlin: Springer series in
statistics.

[7] Arnold, Barry C., \& Strauss, D.  (1991). ``Pseudolikelihood estimation: some examples." \emph{Sankhya: The Indian Journal of Statistics, Series B}: 233-243.

[8] Cox, D.R., \& Hinkley, D.V. (1974). \emph{Theoretical statistics.} Chapman and Hall. ISBN 0-412-12420-3

[9] Lawson, C. L., \& Hanson, R. J. (1974). \emph{Solving least squares problems.} Vol. 161. Englewood Cliffs, NJ: Prentice-hall.

[10] Hong, J., Mohan, K. \& Zeng, D. (2014). ``CVX. jl: A Convex Modeling Environment in Julia."

[11] Domahidi, A., Chu, E., \& Boyd, S. (2013). "ECOS: An SOCP solver for embedded systems." \emph{Control Conference (ECC), 2013 European. IEEE.}

[12] Achanta, R., \& Hastie, T. (2015) "Telugu OCR Framework using Deep Learning." arXiv preprint arXiv:1509.05962 .

\end{document}